# RFBTD: RFB Text Detector

Christen M[1], AB Saravanan[2]

*Signzy.com*

[1]christen@signzy.com, [2]saravanan@signzy.com

*Abstract*— Text detection plays a critical role in the whole procedure of textual information extraction and understanding. On a high note, recent years have seen a surge in the high recall text detectors in scene text images, however text boxes for individual words is still a challenging when dense text is present in the scene. In this work, we propose an elegant solution that promotes prediction of words or text lines of arbitrary orientations and directions, providing emphasis on individual words. We also investigate the effects of Receptive Field Blocks (RFB) and its impact in receptive fields for text segments. Experiments were done on the ICDAR2015 and achieved an F-score of 47.09 at 720p. Implementation: https://github.com/Chris10M/RFB-Text-Detection

I. INTRODUCTION

There has been a recent surge of scene text detectors, after the resurgence of deep neural networks, but one pitfall which is prominent among most of the text detectors is that the maximum size of text instances the detectors can handle is proportional to the receptive field of the network. This encumbers the network to predict longer text regions without increasing the spatial resolution of the input image. Now in order to ameliorate this condition we investigate the effects of receptive field eccentricity [1] and the length of the text lines.

II. RELATED WORK

Scene text detection has been researched tremendously over the years, and has made significant progress. The conventional approaches relied on hand crafted features such as Stroke Width Transform (SWT) [2] and Maximally Stable Extremal Regions (MSER) [3, 4] based methods which generally seek character candidates via edge detection or extremal region extraction. The others [5, 6] also improved the accuracy of conventional detectors but these methods fall way behind the likes of detectors based on Deep Neural Networks (DNN), due to more robust nature and higher accuracy.

The detectors [7, 8, 9, 10, 11, 12, 13, 14, 15, 16, 17, 18] employed deep neural networks in their detection pipeline to significantly improve the accuracy. Xinyu Zhou *et al*. [19] employed a single DNN with a FCN [20] and predicted score map and region proposals. The predictions are then thresholded and Non Maximal Suppression (NMS) is carried out to get the final predicted results.

Songtao Liu *et al*. [21] has investigated the effect of receptive fields and its eccentricity towards positive correlation in object detection performance, their work Receptive Field Block (RFB), has been further explored here. We investigate how the receptive field correlates between having stacked filters of various kernel size and the granularity of the individual words detected across text lines/ segments.

III. METHODOLOGY

The Text detector uses a Resnet [22] backbone, and outputs predictions in the form of rotated boxes, where rotated boxes are represented by 4 channels of axis-aligned bounding boxes and 1 channel for the rotation angle $\theta$. The formulation of the 4 channels represents 4 distances from the pixel location to the top, right, bottom, left boundaries of the rectangle respectively. The basic architecture of the text detector is shown in Fig. 1.

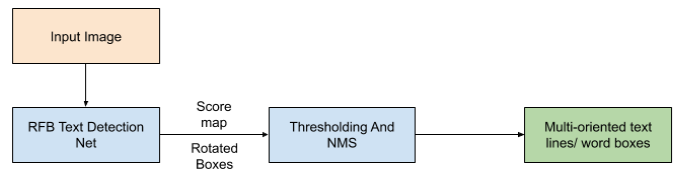

Figure 1. Basic Architecture of RFBTD.

Before discussing the model architecture, we first investigate the role of receptive field in detecting the text regions. In Fig 2. the concentric receptive field enforced by EAST's backbone's symmetrical Conv blocks makes the eccentricity of the receptive field very small concentrated at the center, this creates a need to provide a high resolution feature map across the network to induce maximum region proposals for the given input image.

When the Conv blocks are supplanted by the RFB blocks, which elicits asymmetrical receptive fields by sandwiching filters of various kernel sizes. The RFB block is spatial array of bottlenecks, shortcuts and atrous convolutions as shown in Fig 4. and Fig 5. The two blocks RFB and RFB-s are similar yet have a subtle difference where the 5x5 Conv filters are completely replaced by the 3x3 Conv filters and asymmetrical kernel are also introduced(RFB-s) to reduce the computational overhead of the RFB blocks. These configurations promotes a receptive field of greater eccentricity, as shown in Fig 3.

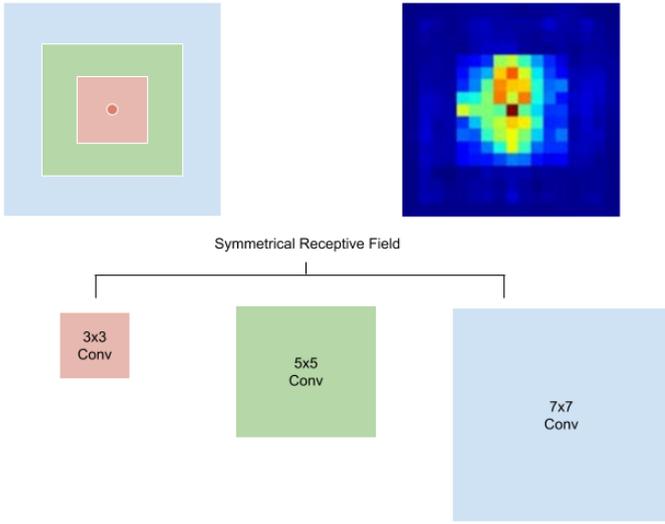

Figure 2. Spatial structures of RFs, in EAST Resnet50 backbone.

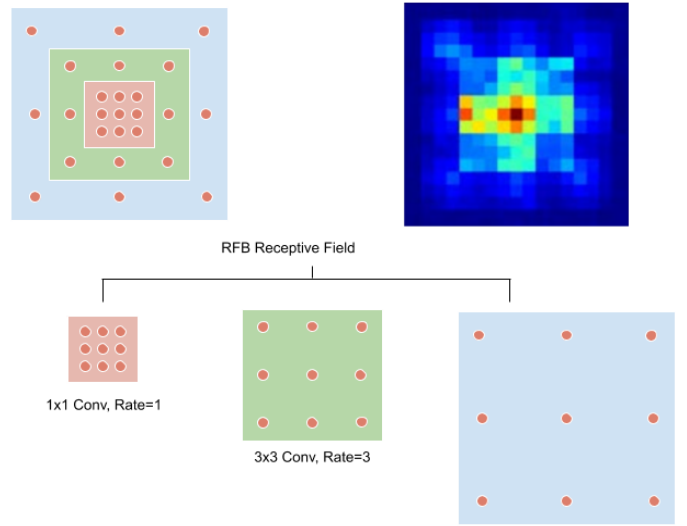

Figure 3. Spatial structures of RFs, in RFBTD

*A. Model Architecture*

The proposed RFB Text Detection (RFBTD) is a fully-convolutional neural(FCN) network which is modelled to output pixel wise predictions of text lines or individual words. The predicted score map and Rotated Box(RBOX) proposals are performed with thresholding and Non Maximal Suppression (NMS) for the final predicted text regions.

The model contains two output branches, one with score map for pixel wise predictions with values in the range of [0, 1]. The other branch outputs RBOX and rotation angle $\theta$ for the proposed regions. The score denotes the confidence of the RBOX proposals in that pixel location. Thresholding is done on the score map to ensure valid proposals, and then NMS is applied for the final RBOXs to get the predicted regions of word boxes/ text lines.

*B. Network*

The network is forked from EAST, but with several modifications. The backbone used is a resnet50, and the feature fusions are done by adding features from the lower layers, rather than concatenating them as shown in Fig 6.

The stem is pre-trained on ImageNet [23] dataset. Five stages of feature maps, denoted $f_i$, are extracted from the stem. For every stage when upsampling we apply 3x3 Conv except $f_5$ to prevent adverse aliasing effects due to the upsampling operation. Each feature map $f_i$ is convolved with a 1x1 Conv block, and then added to the previous layer $f_{i-1}$ feature map, this promotes high spatial resolution of the feature map available at the output of the network. The $f_i$ feature maps are added up to $i=2$ and then the final feature map is produced.

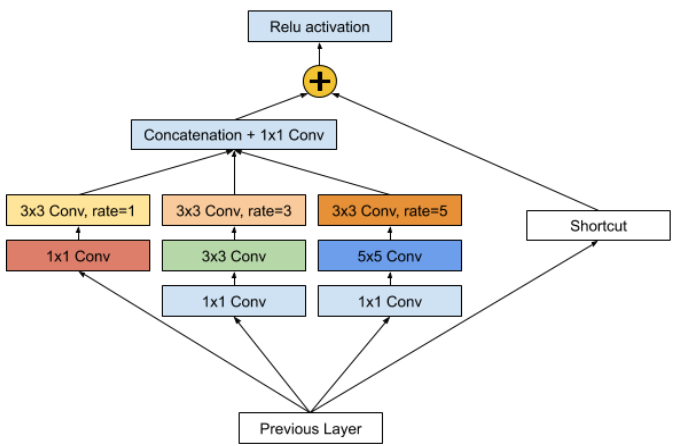

Figure 4. The architecture of RFB

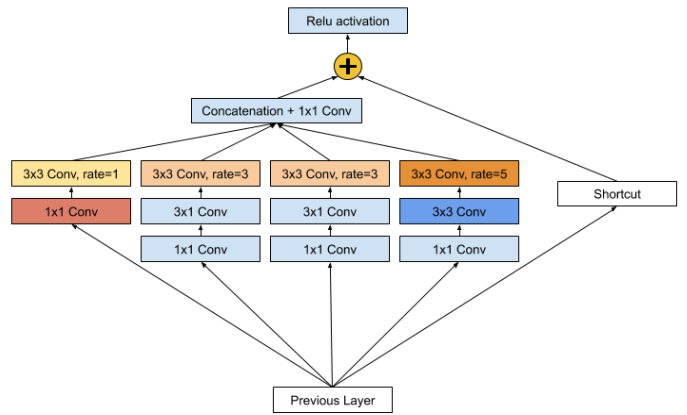

Figure 5. The architecture of RFB-s. The 5x5 Conv layers are replaced with 3 × 3 Conv to reduce the number of parameters.

The final feature map contains two branches from convolving with 1x1 Conv Block, where branch one contains score map *Fs* and a multi-channel map, Rotated boxes which contains four channels of axis-aligned bounding box *R* and 1 channel rotation angle *θ*. The 4 channels represents 4 distances from the pixel location to the top, right, bottom, left boundaries of the rectangle respectively.

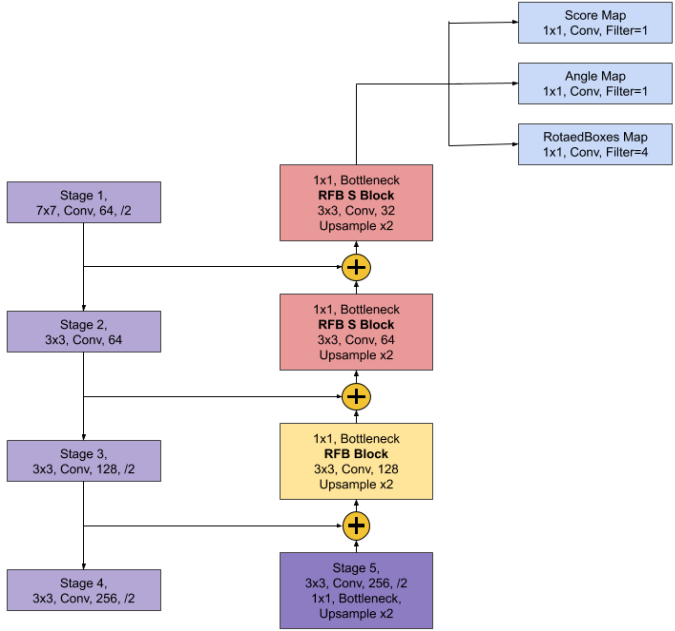

Figure 6. The Network architecture of RFBTD

## C. Loss Functions

$$L = Ls + \lambda g Lg$$

where *Ls* and *Lg* represents the losses for the score map and the RBOX, respectively, and λg weighs the importance between two losses. λg is set to 1.

Dice loss is used for the score map loss *Ls*, as this was tested to be better alternative in faster convergence compared to cross-entropy loss thats is first adopted in text detection by Yao *et al.*

For RBOX loss Lg, we use IOU loss [24], Jiahui Yuet *et al*. The loss favors the intersection area as large as possible even while the predicted box as small as possible, this ameliorates variations in loss due to the huge disparity in the sizes of the text regions / boxes.

Next, the loss of rotation angle is derived from EAST, where the $L_\theta$ computed as

$$L_\theta(\hat{\theta}, \theta^*) = 1 - cos(\hat{\theta} - \theta^*)$$

where $\hat{\theta}$ is the prediction to the rotation angle and $\theta^*$ represents the ground truth. Finally, the overall geometry loss is the weighted sum of RBOX loss and angle loss, given by,

$$Lg = LRBOX + \lambda_\theta L_\theta$$

Where $\lambda_\theta$ is set to 10 in our experiments.

## D. Training

The network is trained end-to-end using ADAGRAD [26] optimizer. To speed up learning, we uniformly sample 512x512 crops from images to form a minibatch of size 16. Exponential decay is induced from one-tenth every 27300 minibatches upto 1e-5. The network is trained until performance stops improving.

## IV. EXPERIMENTS

The proposed method was benchmarked in ICDAR 2015 [25]. It includes a total of 1500 pictures, 1000 of which are used for training and the remaining are for testing. The text regions are annotated by 4 offsets vertices of the quadrangle, by generating RBOX labels by fitting a rotated rectangle which has the minimum area. These images are taken by Google Glass in an incidental way. Therefore text in the scene can be in arbitrary orientations, or suffer from motion blur and low resolution. The RFBTD achieved 47.09 F1-score on this dataset.

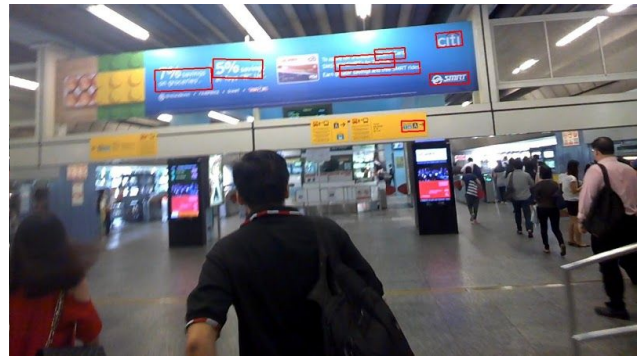

Figure 7. Detector performance of the proposed algorithm on ICDAR 2015.

## V. CONCLUSIONS

A scene text detector that predict word boxes or line level region proposals from arbitrary input images has been proposed. The RFB block module provides an eccentric receptive field which aid in fine granularity to clearly distinguish word boxes in text lines / segments. This helps in many instances, for one it is superior to perform text recognition on words boxes since they offer more accuracy compared to text lines.

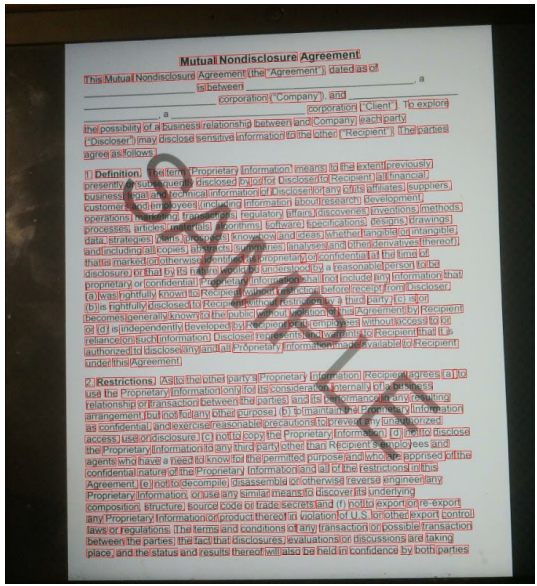

Figure 8a. Detector performance of the proposed algorithm on dense text regions with fine granularity between each text boxes.

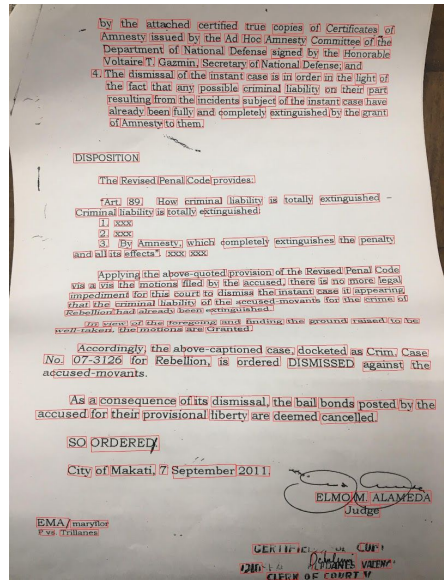

Figure 8b. Detector performance of the proposed algorithm on dense text regions with fine granularity between each text boxes.

## VI. Future Work

The future work can branch out into,
- Invest in a robust detection of curved text.
- Integrating a text recognizer for performing End to End Text spotting.